\documentclass[conference]{IEEEtran}
\IEEEoverridecommandlockouts

\makeatletter
\let\NAT@parse\undefined
\makeatother

\usepackage{hyperref} 
\usepackage{cite}
\usepackage{booktabs} 
\usepackage{multirow}
\usepackage{amsmath,amssymb,amsfonts}
\usepackage{algorithmic}
\usepackage{graphicx}
\usepackage{textcomp}
\usepackage{xcolor}
\usepackage{subcaption}
\usepackage{array}
\usepackage{float}                  
\usepackage{overpic}
\def\BibTeX{{\rm B\kern-.05em{\sc i\kern-.025em b}\kern-.08em
    T\kern-.1667em\lower.7ex\hbox{E}\kern-.125emX}}

\usepackage{graphicx}
\usepackage[ruled]{algorithm2e}
\usepackage{makecell}

\begin{document}

\title{OCGEC: One-class Graph Embedding Classification for DNN Backdoor Detection}

\author{\IEEEauthorblockN{Haoyu Jiang\IEEEauthorrefmark{2},
Haiyang Yu\IEEEauthorrefmark{3},
Nan Li\IEEEauthorrefmark{4} and
Ping Yi\IEEEauthorrefmark{1}\IEEEauthorrefmark{6}}
\IEEEauthorblockA{School of Cyber Science and Engineering,
Shanghai Jiao Tong University, Shanghai, 200240, China\\
\IEEEauthorrefmark{1} Corresponding author\\
Email: \IEEEauthorrefmark{2}jhy549@sjtu.edu.cn,
\IEEEauthorrefmark{3}haiyang yu@sjtu.edu.cn,
\IEEEauthorrefmark{4}lyyqmjshu@sjtu.edu.cn,
\IEEEauthorrefmark{6}yiping@sjtu.edu.cn}}
\maketitle
\begin{abstract} 
Deep Neural Networks (DNNs) have been found vulnerable to backdoor attacks, raising security concerns about their deployment in mission-critical applications. While numerous methods exist to detect backdoor attacks, many of them rely on prior knowledge of the attacks to be detected and require a certain amount of backdoor samples for training, which limits their application in real-world scenarios. This study introduces a novel One-Class Graph Embedding Classification (OCGEC) framework using Graph Neural Networks (GNNs) for model-level backdoor detection. OCGEC first trains a large number of tiny models with a small amount of clean data, then converts these models into graphs to leverage their structural information and weights. A generative self-supervised Graph Auto-Encoder (GAE) is pre-trained on these graphs to learn the representation of DNNs. It is further combined with one-class classification optimization objectives to form a classification boundary between backdoor and benign models, which can effectively detect backdoor models without any knowledge of the attack strategy. Experiments show that our OCGEC achieves AUC scores of more than 98\% against state-of-the-art backdoor attacks on various datasets, outperforming existing backdoor detection methods with a distinctive edge in performance. Note that OCGEC only needs a small amount of clean data and does not rely on any knowledge of the backdoor attacks, making it well-suited for real-world applications.
\end{abstract}





\section{Introduction}

Deep Neural Networks (DNNs) have demonstrated remarkable performance in solving various real-world problems. However, the high cost of training DNNs has led to the rise of third-party online machine learning platforms that provide datasets, computing power, and pre-trained models. Although these platforms offer convenience, they also create opportunities for attackers. The vulnerability of DNNs against backdoor attacks raises serious concerns\cite{li2022backdoor}.

Backdoor attacks can manipulate DNN models by injecting specific triggers into the training dataset or creating a backdoor neural network. Models under backdoor attacks work well on normal inputs. However, when triggered by special inputs, these backdoors grant the attacker complete control over the model's outputs. 



\begin{figure}[t]   
	
	\centering
	
	\includegraphics[width=\linewidth,scale=1]{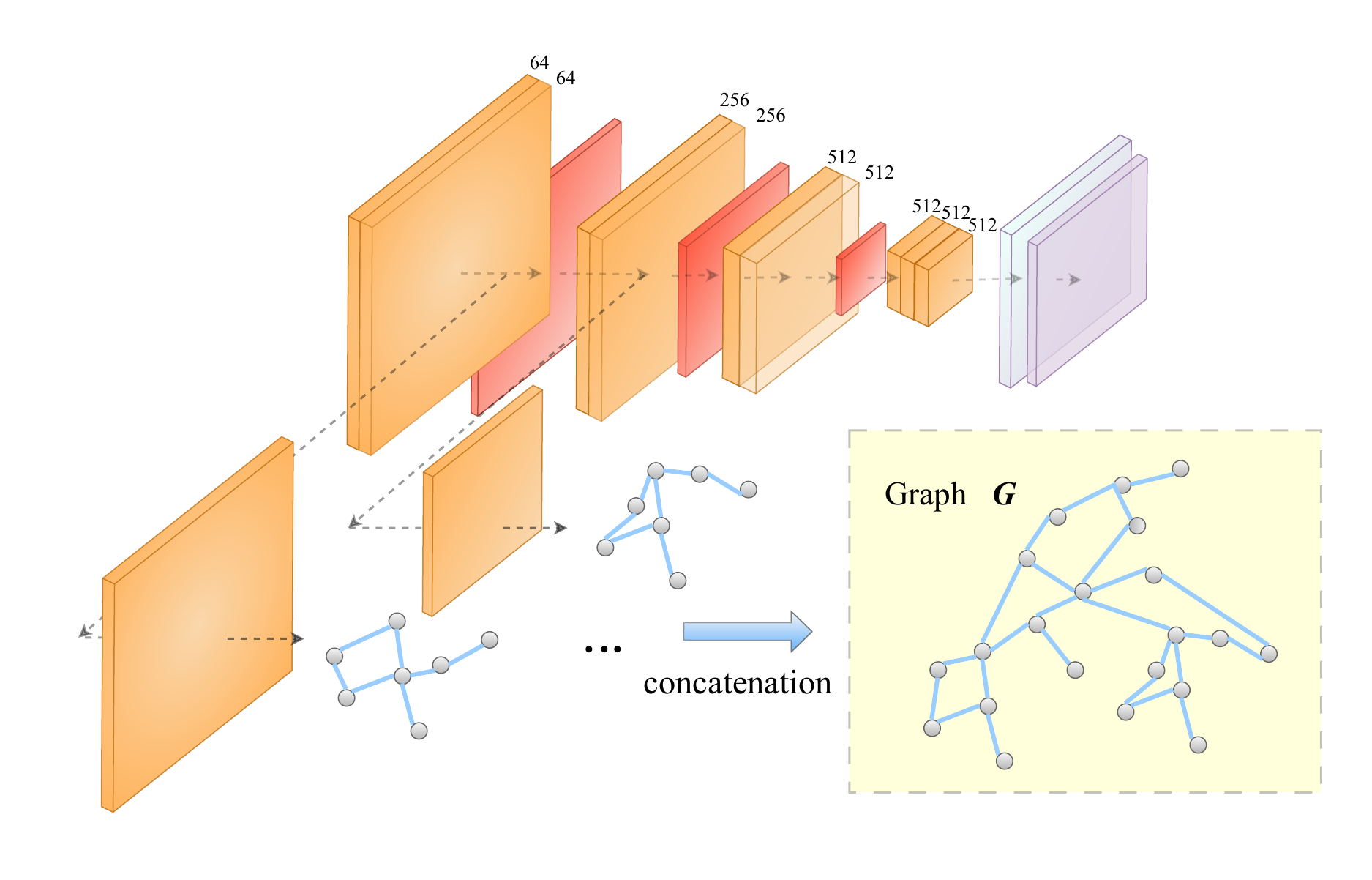}
	
	
	\caption{A novel method for converting model architecture and weights into a graph representation.}
	
	\label{Figuremodeltograph}
	
\end{figure}

There are numerous advanced techniques available for detecting backdoor attacks in DNNs\cite{wang2019neural,gao2019strip,chou2018sentinet,chen2019detecting,chen2019deepinspect,dongBlackboxDetectionBackdoor2021,zheng2022data}. However, most existing detection methods rely on specific assumptions about the attack strategies and full access to the datasets. These dataset-level and input-level methods aim to identify backdoor triggers from poisoned samples or datasets, which can be possibly bypassed by attackers\cite{liu2018trojaning}. 
Besides, the end-users, who need to detect the backdoor, often only have access to the weights
and architectures of the trained models. As a result, the practicality of these methods in real-world scenarios is limited.

It is widely known that the inference or training iterations in DNNs can be represented as computational graphs. For example, MobileNetV2 consists of 17 blocks, each with a similar graph and operation structure. The topology of such blocks can represent their states, enabling us to leverage their redundancy and significance\cite{yu2022topology}. Therefore, we were inspired by such structural information to model a targeted DNN as a graph, as shown in Fig. \ref{Figuremodeltograph}, and explored the potential of GNNs' feature extraction capabilities to differentiate between benign and backdoor models.

In this paper, we propose one-class graph embedding classification (OCGEC), a novel approach for detecting backdoor attacks in DNN models. First, inspired by the computation graph, we create a composition approach that converts all of the target model's weights and structural information into graph data. This method guarantees the complete and accurate preservation of all features in the model. In order to circumvent the constraint of limited access to the poisoned training data, we only use a smaller, clean dataset (i.e., sample data without backdoor triggers) to train tiny models for model-level detection. Training OCGEC with tiny models speeds up the training process, reduces training costs, and lessens OCGEC's dependency on poisoned data.

After the models are converted into graph structure, for non-Euclidean data such as graphs, we utilize the powerful feature extraction capability of GNN to learn their Euclidean data representation. We need to extract valid features from the graphs to distinguish backdoor models from benign models. Therefore, we pre-train a graph auto-encoder that focuses on feature reconstruction using a masking strategy and scaled cosine error. It allows us to obtain the reduced-dimensional model features.

We avoid using poisoned data for backdoor detection training to prevent data imbalance. We modified Deep Support Vector Data
Description (Deep SVDD) \cite{ruff2018deep}to work with GNNs using a few clean data points. Deep SVDD trains the neural network to enclose its outputs within the smallest possible hypersphere, allowing it to capture the common features that represent variations in data distribution. This method enables us to train an effective one-class classifier that forms a tight boundary around the benign models, allowing for the detection of backdoor attacks.

Our method is applicable to diverse attacks and application domains since OCGEC makes no assumptions about the attack strategy and only needs a few clean samples. Experimental results show that our OCGEC achieves excellent performance in detecting backdoor models against various backdoor attacks across diverse datasets of multiple modals, including image, audio, and text, surpassing the detection performance of state-of-the-art backdoor detection techniques. This validates the effectiveness of our model-to-graph and graph embedding methods. Moreover, we present evidence that the trained OCGEC exhibits strong generalization capabilities in identifying previously unseen backdoors. Our contributions can be summarized as follows:
\raggedbottom
\begin{itemize}
\item [$\bullet$]
We propose a novel OCGEC framework leveraging the representation ability of GNNs for backdoor detection tasks. To the best of our knowledge, this is one of the first GNN-based projects in this scenario.
\end{itemize}
\begin{itemize}
\item [$\bullet$]
We develop a novel model-to-graph approach that can efficiently capture the structural information and weight features of the DNNs, which proves highly effective for backdoor detection.
\end{itemize}

\begin{itemize}
    \item [$\bullet$]
We empirically show the effectiveness and generalizability of our approach through extensive evaluation and comparison with popular backdoor detection approaches against various types of backdoor attacks on diverse datasets. 
\end{itemize}

\section{Related Work}
\subsection{Backdoor Attack}

Attackers may create neural network models with hidden backdoors with different attack strategies that operate normally under normal circumstances but behave maliciously when presented with manipulated inputs. 

Subsequent developments in attack strategies include input-space attacks and feature-space attacks. For instance, the Modification Attack, also introduced by Gu et al.\cite{guBadNetsIdentifyingVulnerabilities2019}, involves selectively altering training samples with a trigger pattern and re-inserting them into the training set with malicious labels. Chen's Blending Attack \cite{chen2017targeted} represents another approach, where the attacker subtly merges a trigger, like background noise, into the original input. A more advanced technique, presented by Xu et al. \cite{xu2021detecting}, employs distributed algorithms for imperceptible global image updates, proving robust against many defense strategies. Furthermore, to counter reverse engineering of triggers, attackers have developed dynamic strategies like the WaNet \cite{nguyen2020wanet}. This paper will focus on implementing these poisoning-based backdoor attacks in experimental settings.

\subsection{Backdoor Detection}
Backdoor defense involves two primary tasks: backdoor detection and removal. Detection identifies backdoored models or poisoned datasets, whereas removal removes the injected backdoor from the infected model while minimizing performance damage on clean samples. 

Backdoor detection can implemented at the input, dataset, or model level. The input level detection \cite{gao2019strip,tran2018spectral,chou2018sentinet,sarkar2020backdoor,fu2020detecting}is designed to include triggers at network inputs to prevent backdoor activation. On the other hand, assuming that the detector has access to the dataset, the backdoor can be identified with the statistical deviation in the feature space by dataset-level detection \cite{chen2019detecting,xiang2019benchmark,peri2020deep}. In addition, model-level detection such as \cite{liuABSScanningNeural2019,xu2021detecting,kolouri2020universal} determines whether a model contains backdoor triggers directly on a model-by-model basis, and this type of approach is more practical. Existing methods typically require training data access, neural network architectures, types of triggers, target classes, etc. Our OCGEC, however, is capable of overcoming these issues.

\subsection{Graph Neural Networks }
Graph neural networks (GNNs) have developed as a powerful method for processing non-Euclidean data, excelling in tasks like node-level classification, link-level prediction, and graph-level classification\cite{wuComprehensiveSurveyGraph2021} through a neighborhood aggregation approach. Within the graph auto-encoder (GAE) framework, GNNs are particularly effective for graph embedding and classification\cite{tianHeterogeneousGraphMasked2023}, outperforming traditional graph kernel methods. This supports our model-to-graph feature extraction approach. In this paper, We use a GAE for generative self-supervised graph pre-training with a masking strategy like Hou et al.\cite{houGraphMAESelfSupervisedMasked2022}.

\begin{figure*}[t]   
	
	\centering
	
	\includegraphics[scale=0.43]{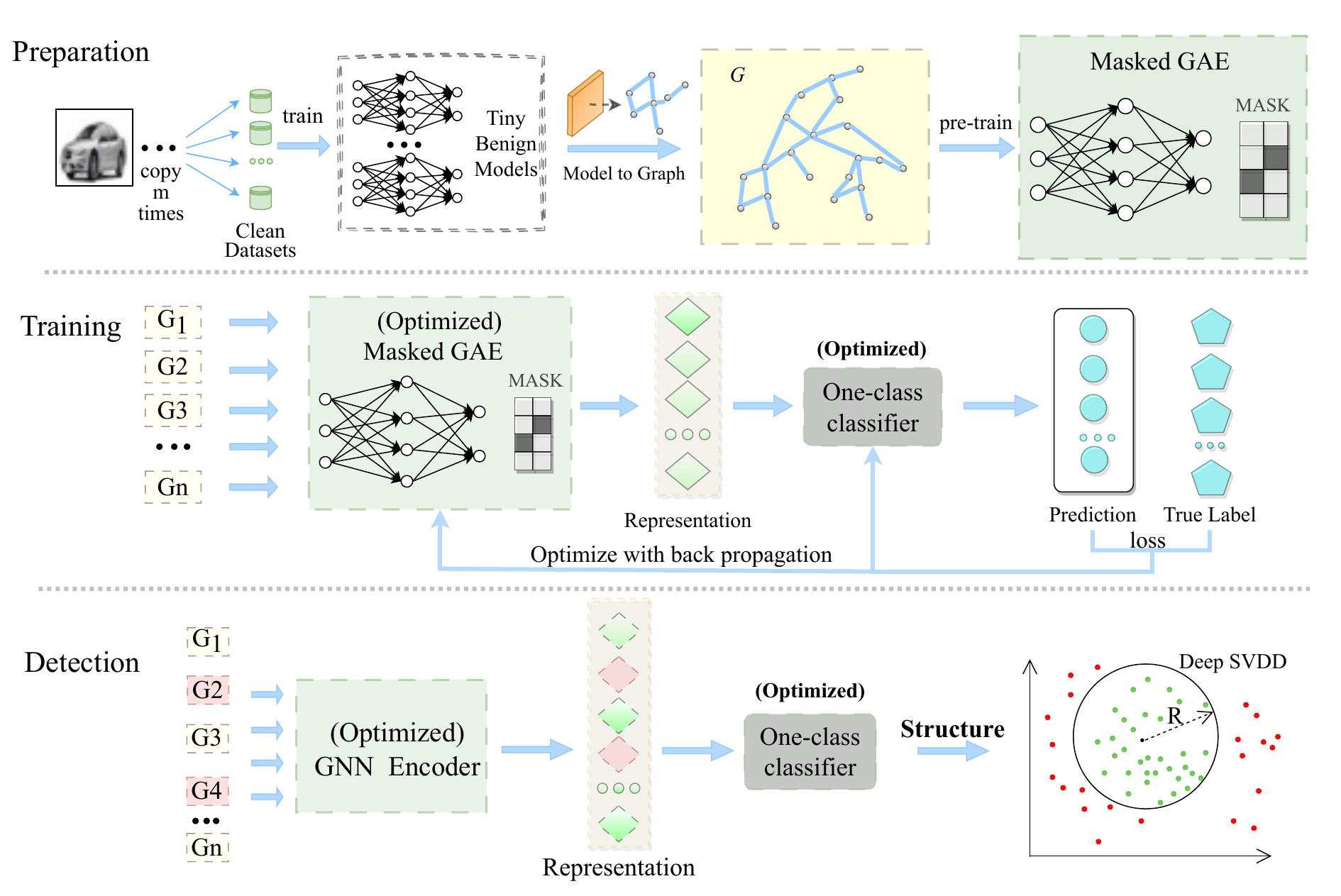}
	
	
	\caption{Overview of OCGEC framework, including graph preparation, classifier training, and backdoor detection phases.}
	
	\label{OCGEC}
	
\end{figure*}

\subsection{Applications of One-class Classification}

One-class classification is often employed in scenarios with extreme data imbalance to ascertain if a target is part of the class learned in training. Recent studies \cite{chandola2009anomaly,khanOneclassClassificationTaxonomy2014,pang2021deep,pimentel2014review} have proposed various one-class techniques. However, most of these techniques are designed for low-dimensional data inputs and struggle with deep feature extraction, especially in image datasets \cite{chandola2009anomaly,khanOneclassClassificationTaxonomy2014}. In the past few years, deep learning-based one-class methods\cite{pang2021deep,pimentel2014review} have emerged. These methods utilize loss functions inspired by statistical methods such as OC-SVM \cite{manevitz2001one} and SVDD \cite{tax2004support} or utilize regularization techniques to make traditional neural network training more suitable for one-class classification. Our paper adopts one-class classification because it is challenging to obtain a sufficient quantity of poisoned data for training in backdoor detection scenarios.

	
	
	
	
	

\section{Methodology}
Our proposed OCGEC is illustrated in Figure \ref{OCGEC}. OCGEC works with the target DNNs to be detected and a small subset of clean data. It can be divided into three phases. In the following, we will explain the details of graph preparation, classifier training, and backdoor detection.




\subsection{Threat Model}
Following previous work\cite{xu2021detecting}, we assume that the adversary has full access to the training dataset and can employ arbitrary attack methods to generate backdoor models. Trigger patterns can take on any shape, position, and size. We assume that the defender has access to the model's weights and architectures but possesses only a small number of clean samples.

\subsection{Graph Preparation}\label{Graph Preparation}
OCGEC aims to identify backdoor models directly by their weights and architecture. Since we only get a small set of clean data, to capture the information of benign models, we consider training a set of tiny benign models on the clean data and using their weights to train our method. To ensure the diversity of weight features and enhance the robustness of our method, we train these tiny benign models under different hyper-parameter settings, including parameter initialization methods, learning rates, and training epochs. Then, the tiny benign models are converted into graphs based on their structure and weights.

Consider a Convolutional Neural Network (CNN) consisting of $n$ convolutional layers $(L_{1}, L_{2}, \ldots,L_{n})$. Batch normalization and non-linear activation functions are omitted here for simplicity. The network can be expressed as a function of input $x$ and parameter $\theta$:
\begin{equation}
    f(x; \theta) = L_n\circ L_{n-1}\circ \ldots \circ L_2\circ L_1
\end{equation}

Consider a graph $\mathcal{G} = (\mathcal{V}, \mathcal{E}, \mathbf{X})$ constructed for the CNN $f$ with $n$ convolutional layers and $N$ filters in total, where $\mathcal{V}$, $\mathcal{E}$ and $\mathbf{X}\in \mathbb{R}^{N\times d}$ denote the node set, the edge set, and the input node feature matrix, respectively. Note that $d$ represents the input node feature size. Each node in $\mathcal{V}$ corresponds to a specific filter in the corresponding convolutional layer of $f$. Hence we can divide the node set $\mathcal{V}$ into $n$ subsets $(\mathcal{V}_1, \mathcal{V}_2, \ldots, \mathcal{V}_n)$ by the layer that each node corresponds to. As the filters in consecutive layers are fully connected, the edge set $\mathcal{E}$ can be determined as:
\begin{equation}
    \mathcal{E}=\{(v_i, v_j)|v_i\in \mathcal{V}_t, v_j\in \mathcal{V}_{t+1}, t=1,2,\ldots,n-1\},
\end{equation}
making $\mathcal{G}$ a n-partite graph. For each node, we consider the weight matrix $\theta^{c\times w\times h}$ of the corresponding filter as its feature, where $c$, $w$, $h$ represent the number of channels, the width, and the height of the convolutional kernel, respectively. The input feature matrix $\mathbf{X}$ is directly the concatenation of $N$ weight matrices, with the input feature size $d = c\times w\times h$.

The graph-structured data preserves the complex structural information and weight features of DNNs. With the help of GNNs, we can hierarchically extract graph embeddings by passing, transforming, and aggregating information between nodes.  The experimental results in Section \ref{exp} further reveal the efficiency of our model-to-graph methods in capturing intricate deep features of DNNs.

\subsection{Pre-training Graph Auto-Encoder}\label{GAE}
Due to the sparsity of DNNs, the graph constructed through our model-to-graph method contains a large number of nodes and complex structural information. We consider training a GAE through self-supervised learning to extract essential information from large-scale graphs. We denote $\mathbb{E}$ as the graph encoder and $\mathbb{D}$ as the graph decoder, with $\mathbf{W}_E$ and $\mathbf{W}_D$ as their weights. $\mathbb{E}$ and $\mathbb{D}$ are both Graph Isomorphism Networks (GINs) since GIN is well-suited to capture graph-level information.

Following previous work \cite{houGraphMAESelfSupervisedMasked2022}, we apply a mask-based training strategy and focus our GAE on the task of feature reconstruction. Given a graph $\mathcal{G} = (\mathcal{V}, \mathcal{E}, \mathbf{X})$, the output of $\mathbb{E}$ is a representation matrix of nodes after applying a masking strategy:
\begin{equation}
\label{eq9}
    \mathbf{H}_{mask}= \mathbb{E}(\mathcal{E},\text{M}(\mathbf{X}), \mathbf{W}_E),
\end{equation}
where $\text{M}$ is the random mask function that randomly selects a certain proportion of graph nodes and sets their features as all-zero vectors. The decoder maps the representation to reconstruct the input features of nodes:
\begin{equation}
\mathbf{X}^{\prime}=\mathbb{D}(\mathcal{E},\mathbf{H}_{mask}, \mathbf{W}_D)
\end{equation}

The objective of pre-training is to minimize the scaled cosine error (SCE) loss of the reconstructed feature matrices on a set of graphs constructed for the tiny models:

\begin{equation}\label{equ10}
    \min_{\substack{\mathbf{W}_E, \mathbf{W}_D}} \sum_{(\mathcal{V}, \mathcal{E}, \mathbf{X})}\text{SCE}(\mathbf{X}, \mathbb{D}(\mathcal{E}, \mathbb{E}(\mathcal{E}, M(\mathbf{X}), \mathbf{W}_E), \mathbf{W}_D))
\end{equation}

By pre-training the GAE with the objective function defined in Equation \ref{equ10}, we can get a robust graph encoder $\mathbb{E}$ with weight $\mathbf{W}_E$, which has learned meaningful representations of DNNs and can efficiently capture crucial features of the graph data.


\subsection{One-class Learning}
In this section, we combine the graph encoder $\mathbb{E}$ and the Deep SVDD algorithm to obtain our OCGEC. Instead of directly using the pre-trained weight $\mathbf{W}_E$, we consider jointly optimizing the weight of $\mathbb{E}$ and minimizing the volume of the data-enclosing hypersphere learned by Deep SVDD to get a more insightful representation space.

\noindent\textbf{Hierarchical Graph Embedding. }Different from typical graph-structured data like social networks or chemical molecules, the computational graph of DNNs has a special hierarchical structure, which inspires us to get graph embeddings from node representations in a hierarchical way. Given the node representation matrix $\mathbf{H}$ and the node set $\mathcal{V} = (\mathcal{V}_1, \mathcal{V}_2, \ldots, \mathcal{V}_n)$ divided into $n$ partites, let $h_j$ denote the $j$-th row of $\mathbf{H}$, i.e., the representation of the $j$-th node, the graph embedding function $\text{Emb}(\cdot)$ can be defined as follows:
\begin{equation}
\text{Emb}(H, \mathcal{V})=\underset{i=1} {\overset{n} {\Big \Vert}}(\operatorname{MeanPooling}(\{h_{j} \mid v_j \in \mathcal{V}_i\}),
\end{equation}
where $\Vert$ denotes the concatenation operation, and $\operatorname{MeanPooling}$ denotes the average pooling for nodes in each partite.

\noindent\textbf{Objective of OCGEC. }\ OCGEC leverages \textit{Deep SVDD} \cite{pereraOneClassClassificationSurvey2021} to learn a concise hypersphere boundary with center $\boldsymbol{c}$ in representation space $\mathcal{F}_{k}$ and radius $R>0$ that encompasses all the training data, i.e., benign models. The hypersphere is a descriptive boundary for benign models and can be used to identify outliers, i.e., backdoored models. Given a set of graphs $(\mathcal{G}_1, \mathcal{G}_2, \ldots, \mathcal{G}_k)$ with $\mathcal{G}_i = (\mathcal{V}_i, \mathcal{E}_i, \mathcal{X}_i)$ the objective of OCGEC is to minimize the volume of the data-enclosing hypersphere with the graphs:
\begin{equation}
\begin{aligned}
\min _{\substack{\mathbf{W}_E,  \boldsymbol{c}, R}} &R^2+\frac{1}{\nu k}\sum_{i=1}^k \max \{0, \Vert \text{Emb}\left(\mathbb{E}(\mathcal{E}_i,\mathbf{X}_i, \mathbf{W}_E))-\boldsymbol{c}\right\Vert^2 \\
&-R^2\} +\frac{\lambda}{2}\|\mathbf{W}_E\|^2,
\end{aligned}
\label{equ14}
\end{equation}
where the first term $R^2$ is to minimize the volume of the hypersphere, the second term is to penalize the distance between the graph embeddings and the center $\boldsymbol{c}$, hyperparameter $\nu \in (0; 1]$ controls the trade-off between the volume of the sphere the violations of the boundary, and the last term with the hyperparameter $\lambda$ is a weight decay regularizer. 

\noindent\textbf{Optimization of OCGEC. } \  As is specified in Equation \ref{equ14}, the optimization objective can be characterized by three parameters, namely: weights of the graph encoder $\mathcal{W}_E$, hypersphere radius $R$ and hypersphere center $\boldsymbol{c}$. We optimize the parameter $\mathcal{W}$ of the graph encoder through back-propagation. Details of the optimization process are summarized in Algorithm \ref{alg:Framwork}.
\begin{algorithm}[htb] 
\caption{ Training phase of OCGEC} 
\label{alg:Framwork} 
\KwIn{ $k$ graphs generated by tiny models $\mathcal{G}_i = (\mathcal{V}_i,\mathcal{E}_i,\mathbf{X}_i)$,
weight decay $\lambda \textgreater 0 $, slack hyperparameter $\nu\in (0; 1]$;}
\KwOut{
weights $\mathbf{W}_E$, center $\boldsymbol{c}$, radius $R$ ;}
Initialize $\mathcal{W}_E$ with the pre-trained weights of $\mathbb{E}$\; 
Initialize $R=0$, $c =\frac{1}{k} \sum_{i=1}^k\text{Emb}(\mathbb{E}(\mathcal{E}_i,\mathbf{X}_i, \mathbf{W}_E))$\; 
\While{\rm{epoch} $\textless$ \rm{max epoch budget}}{
    {\small$d_i = \Vert\text{Emb}\left(\mathbb{E}(\mathcal{E}_i,\mathbf{X}_i, \mathbf{W}_E))-\boldsymbol{c}\right\Vert^2;$ \\}
    {\small $\mathcal{L}=R^2+\frac{1}{\nu k} \sum_{i=1}^N \max\{0, d_i-R^2\} +\frac{\lambda}{2}\|{\mathbf{W}_E}\|^2;$\\}
    {\small Update $\mathbf{W}_E$ by its stochastic gradient $\nabla _{\mathbf{W}_E}(\mathcal{L})$;}\\
    {$c =\frac{1}{k} \sum_{i=1}^k\text{Emb}(\mathbb{E}(\mathcal{E}_i,\mathbf{X}_i, \mathbf{W}_E))$}\\
    {\small Update $R$ using $(1-\nu)\times100\%$ percentile of $\{d_i\}^k$}
}
\textbf{return} $\mathbf{W}_E$, $\boldsymbol{c}$ and $R$ ; 
\end{algorithm}

\noindent\textbf{Backdoor Detection. } After the target models are converted into graphs and the graph embeddings are obtained, graph embeddings that fall outside the hypersphere defined by $\boldsymbol{c}$ and $R$, specifically those satisfying $\Vert\text{Emb}\left(\mathbb{E}(\mathcal{E},\mathbf{X}, \mathbf{W}_E))-\boldsymbol{c}\right\Vert^2 \textgreater R^2$, are classified as outliers and thus identified as backdoor models. Benign models, conversely, reside within the boundary of the hypersphere.

\section{Experiments}\label{exp}
\subsection{Experimental Setup}
\subsubsection{Datasets}
We evaluate three typical machine learning tasks: vision, speech, and natural language processing. For the vision task, we use the classical GTSRB\cite{stallkamp2011german} and CIFAR10\cite{krizhevsky2009learning} datasets.
For the speech task, we use the Speech Command v2 dataset\cite{warden2018speech}.
For the NLP dataset, we use MR Movie Reviews \cite{kim2014convolutional}, which is a corpus of movie reviews used for sentiment analysis. For all tasks in this paper, we use only 2\% of the clean dataset to generate 2048 benign tiny models for training and use 128  benign tiny models and the other 128 backdoor tiny models with a backdoor dataset generated from 50\% of the dataset for testing.


\subsubsection{Attack Settings}
\begin{figure}[t]
  \centering
  \begin{subfigure}[b]{0.09\textwidth}
    \includegraphics[width=\textwidth]{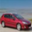}
    \label{fig:sub1}
  \end{subfigure}
    \vspace{-3mm}
  \begin{subfigure}[b]{0.09\textwidth}
    \includegraphics[width=\textwidth]{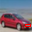}
    \label{fig:sub2}
  \end{subfigure}
  \begin{subfigure}[b]{0.09\textwidth}
    \includegraphics[width=\textwidth]{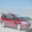}
    \label{fig:sub3}
  \end{subfigure}
  \begin{subfigure}[b]{0.09\textwidth}
    \includegraphics[width=\textwidth]{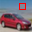}
    \label{fig:sub4}
  \end{subfigure}
  \begin{subfigure}[b]{0.09\textwidth}
    \includegraphics[width=\textwidth]{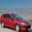}
    \label{fig:sub5}
  \end{subfigure}

  \begin{subfigure}[b]{0.09\textwidth}
    \includegraphics[width=\textwidth]{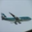}
    \caption{original}
    \label{fig:sub6}
  \end{subfigure}
  \begin{subfigure}[b]{0.09\textwidth}
    \includegraphics[width=\textwidth]{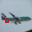}
    \caption{-M}
    \label{fig:sub7}
  \end{subfigure}
  \begin{subfigure}[b]{0.09\textwidth}
    \includegraphics[width=\textwidth]{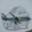}
    \caption{-B}
    \label{fig:sub8}
  \end{subfigure}
  \begin{subfigure}[b]{0.09\textwidth}
    \includegraphics[width=\textwidth]{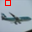}
    \caption{-J}
    \label{fig:sub9}
  \end{subfigure}
  \begin{subfigure}[b]{0.09\textwidth}
    \includegraphics[width=\textwidth]{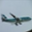}
    \caption{-W}
    \label{fig:sub10}
  \end{subfigure}
  \caption{Attacks types on CIFAR10 dataset: -M for Modification, -B for Blended, -J for Jumbo, and -W for WatNet.}
  \label{fig:figure}
\end{figure}

We evaluate our OCGEC against various well-known attack methods, including Modification  Attack\cite{guBadNetsIdentifyingVulnerabilities2019}, Blending Attack\cite{chen2017targeted}, Jumbo Attack\cite{xu2021detecting}and WaNet\cite{nguyen2020wanet}. Additionally, we consider two white-box adaptive attacks. For each dataset except MR, we generate 128 backdoored tiny models using these attacks. Only the modification attack is applied to the discrete MR dataset. We test the Modification Attack with All-to-One and All-to-All strategies. The effects of the attack are illustrated in Fig.\ref{fig:figure}.

\subsubsection{OCGEC Setup}
To demonstrate the OCGEC framework's empirical performance, we standardized training across all datasets and paradigms using consistent hyperparameters. CNNs modeled the GTSRB, CIFAR10, and MR datasets, while SC employed RNNs, with an architecture extension incorporating ResNet, DenseNet, and others for CIFAR10. We opted for the improved Adam optimizer\cite{zhang2018improved} with a learning rate of 0.001. During pre-training, a maximum epoch of 50 was set for the graph auto-encoder's initial training. A two-layer GIN, with sizes 64-32 for GTSRB and CIFAR10 (CNNs) and 128-64 for CIFAR10 (ResNet-18) and MR, was utilized. SC's model had a size of 1024-512. Feature dropout ratio and masking rate were set to 0.2 and 0.75, respectively, with $MeanPooling$ for pooling. OCGEC applied weight decay (0.0005 coefficient) and an early stopping mechanism based on OCGEC loss[Eq.\ref{equ14}] and test set AUC, with a maximum of 10 epochs and a patience of 2 epochs.


\subsubsection{Baselines}
In this paper, we choose six existing works on backdoor attack detection as our baselines: Activation Clustering (AC)\cite{chen2019detecting}, Neural Cleanse (NC)\cite{wang2019neural}, Spectral Signature (Spectral)\cite{tran2018spectral}, STRIP\cite{gao2019strip}, MNTD-OCC\cite{xu2021detecting} and ABD\cite{zhangAdaptiveBlackboxBackdoor2022}. We reconstruct the AC, Spectral, and STRIP implementations with PyTorch to accommodate the model-level detection. AC detects backdoors in datasets by identifying inconsistencies in network activation using the ExRe score\cite{chen2019detecting}. Spectral measures the outlier value of the input, STRIP perturbs the input to test for backdoors, and NC explores model-level detection. MNTD-OCC uses query tuning and the OCSVM algorithm for black-box detection. These methods, except MNTD-OCC, are trained for binary classification using equal numbers of benign and backdoor models, which can be difficult in real-world scenarios. Despite this, our detection performance is superior to theirs.


\subsection{Main Results}
\subsubsection{Backdoor Attack Performance}  To measure the performance of backdoor tiny models and the effectiveness of backdoor attacks, we calculate the classification accuracy and attack success rate of the test set on different tiny backdoor models. Also, as a comparison, we calculate the classification accuracy of the benign tiny models. The results are shown in Table \ref{ASR}. The benign tiny models converted to graphs have limited classification accuracy due to the fact that they are trained using only a very small amount of clean dataset (2\%), at which point the training size is very small. Using 50\% of the dataset to generate the backdoor tiny models, the accuracy of the tiny model's classification is significantly improved and makes the attack success rate close to 100\% as well. 
\begin{table}[h]
  \centering
  \caption{The classification accuracy and attack success rate on four datasets.}
  \scalebox{0.83}{
    \begin{tabular}{cccc}
    \toprule
    \multicolumn{1}{c}{\multirow{2}{*}{Models}} & \multicolumn{1}{c}{Tiny Benign Models (2\%)} & \multicolumn{2}{c}{Tiny Backdoor Model (50\%)} \\
            
        \cmidrule(lr){2-2} \cmidrule{3-3} \cmidrule{4-4}
          & Accuracy   & Accuracy  & \multicolumn{1}{l}{Attack Success Rate} \\
    \midrule
    GTSRB-benign & 44.61±0.03   & 97.37±0.02 & - \\
    GTSRB-M & -          & 97.15±0.02 & 99.78±0.01 \\
    GTSRB-B & -          & 97.24±0.03 & 99.62±0.04 \\
    GTSRB-J & -          & 96.63±0.75 & 99.62±0.04 \\
    GTSRB-W & -          & 97.51±0.01 & 100 \\
    CIFAR 10-benign  & 42.31±0.06      & 61.24±0.04 & - \\
    CIFAR 10-M & -          & 60.94±0.08 & 99.65±0.03\\
    CIFAR 10-B & -          & 59.32±0.06 & 89.52±0.20 \\
    CIFAR 10-J & -          & 60.25±0.50 & 96.78±0.20 \\
    CIFAR 10-W & -          & 61.46±0.05 & 100 \\
    SC-benign & 68.25±0.02      & 83.46±0.02 & -\\
    SC-M  & -          & 83.13±0.03 & 98.66±0.02 \\
    SC-B  & -          & 82.20±0.08 & 98.82±0.02\\
    SC-J  & -          & 82.44±0.10 & 97.67±0.20\\
    SC-W  & -          & 82.83±0.05 & 100 \\
    MR-benign & 70.62±0.01      & 74.36±0.02 & -\\
    MR-M  & -         & 74.48±0.01 & 97.42±0.05\\
    MR-B  & \textbackslash{}         & \textbackslash{} & \textbackslash{} \\
    MR-J  & -         & 73.62±0.75 & 98.65±0.33\\
    MR-W  & -          & 74.69±0.02 & 100 \\
    \bottomrule
    \end{tabular}}
  \label{ASR}%
\end{table}%
\begin{table*}[t]
  \centering
  \caption{Detection AUC for each method in \%.}
  \scalebox{1}{
    \begin{tabular}{cccccccccccccccc}
    \toprule
    \multirow{2}[0]{*}{Methods} & \multicolumn{4}{c}{GTSRB}     & \multicolumn{4}{c}{CIFAR10}   & \multicolumn{4}{c}{SC}        & \multicolumn{3}{c}{MR} \\
    \cmidrule(lr){2-5} \cmidrule(lr){6-9}  \cmidrule(lr){10-13} \cmidrule(lr){14-16}  
          &  -M   &  -B   &  -J   &  -W   &  -M   &  -B   &  -J   &  -W   &  -M   &  -B   &  -J   &  -W   &  -M     &  -J   &  -W \\
    \midrule
    AC\cite{chen2019detecting}    & 73.27 & 78.61 & 84.76 & 69.53 & 85.94 & 74.61 & 72.65 & 65.23 & 79.69 & 82.81 & 83.59 & 72.26 & 87.89  & 85.54 & $\leq$50 \\
    
    NC\cite{wang2019neural}    & 91.4  & \textbf{89.84} & 92.57 & 82.81 & 53.91 & 57.42 & $\leq$50   & $\leq$50   & 91.41 & \textbf{96.48} & 95.71 & 84.38 & \textbackslash{}  & \textbackslash{} & \textbackslash{} \\
    
    Spectral\cite{tran2018spectral} & $\leq$50   & 51.17 & $\leq$50   & $\leq$50   & 88.28 & 56.64 & 61.71 & $\leq$50   & $\leq$50   & $\leq$50   & $\leq$50   & $\leq$50   & \textbf{95.31}  & 92.57 & 88.28 \\
    
    STRIP\cite{gao2019strip} & 83.59 & 66.41 & 78.13 & 92.58 & 85.55 & 81.64 & 80.47 & 89.84 & 86.97 & 85.16 & 89.45 & 71.48 & \textbackslash{}  & \textbackslash{} & \textbackslash{} \\

    MNTD-OCC\cite{xu2021detecting} & 76.17 & 73.43 & 73.83 & $\leq$50   & 65.23 & 76.56 & 72.66 & $\leq$50   & 88.67 & 85.94 & 87.11 & 81.25 & $\leq$50   & 52.73 & $\leq$50 \\
    ABD\cite{zhangAdaptiveBlackboxBackdoor2022} & 94.14 & 85.94 & 87.89 & 71.09 & 90.23 & \textbf{88.28} & 87.50& 90.23 & 65.63 & 87.12 & 78.91& 82.92&\textbackslash{}  & \textbackslash{}  & \textbackslash{}  \\

\textbf{OCGEC} & \textbf{97.27} & 88.19 & \textbf{99.09} & \textbf{99.61} & \textbf{93.36} & 85.94 & \textbf{98.42} & \textbf{100}   & \textbf{94.53} & 91.41 & \textbf{98.23}  & \textbf{96.50}  & 93.75  & \textbf{92.97} & \textbf{100} \\
    \bottomrule
    \end{tabular}}%
  \label{tab:results}%
\end{table*}%
\begin{table}[h]
  \centering
  \caption{The detection AUC of each approach against All-to-All Attack on GTSRB and CIFAR10.}
    \begin{tabular}{ccc}
    \toprule
    Methods & \multicolumn{1}{l}{GTSRB-ATA} & \multicolumn{1}{l}{CIFAR10-ATA} \\
    \midrule
    AC    & 90.37 & 77.41 \\
    NC    & 51.36 & 52.44 \\
    Spectral & 84.36 & $\leq$50 \\
    STRIP & 61.60 & $\leq$50 \\
    MNTD-OCC & 97.29 & 70.18 \\
    ABD   & 87.45 & 73.77 \\
\textbf{OCGEC} & \textbf{98.74} & \textbf{92.38} \\
    \bottomrule
    \end{tabular}%
  \label{AlltoAll}%
\end{table}%

\subsubsection{Backdoor Detection Performance}  Table \ref{tab:results} shows the detection performance with Area
Under ROC Curve (AUC) as the metric. All of the baselines we compare our method with make certain assumptions about backdoor attacks. None of these methods demonstrate universal effectiveness, as they are primarily designed for image-related tasks and lack support for tasks involving speech or text. For example, NC, STRIP, and ABD can't be extended to NLP tasks.  

\begin{figure}[t]   
	
\centering
	
\begin{subfigure}[b]{0.22\textwidth}
    \includegraphics[width=\textwidth]{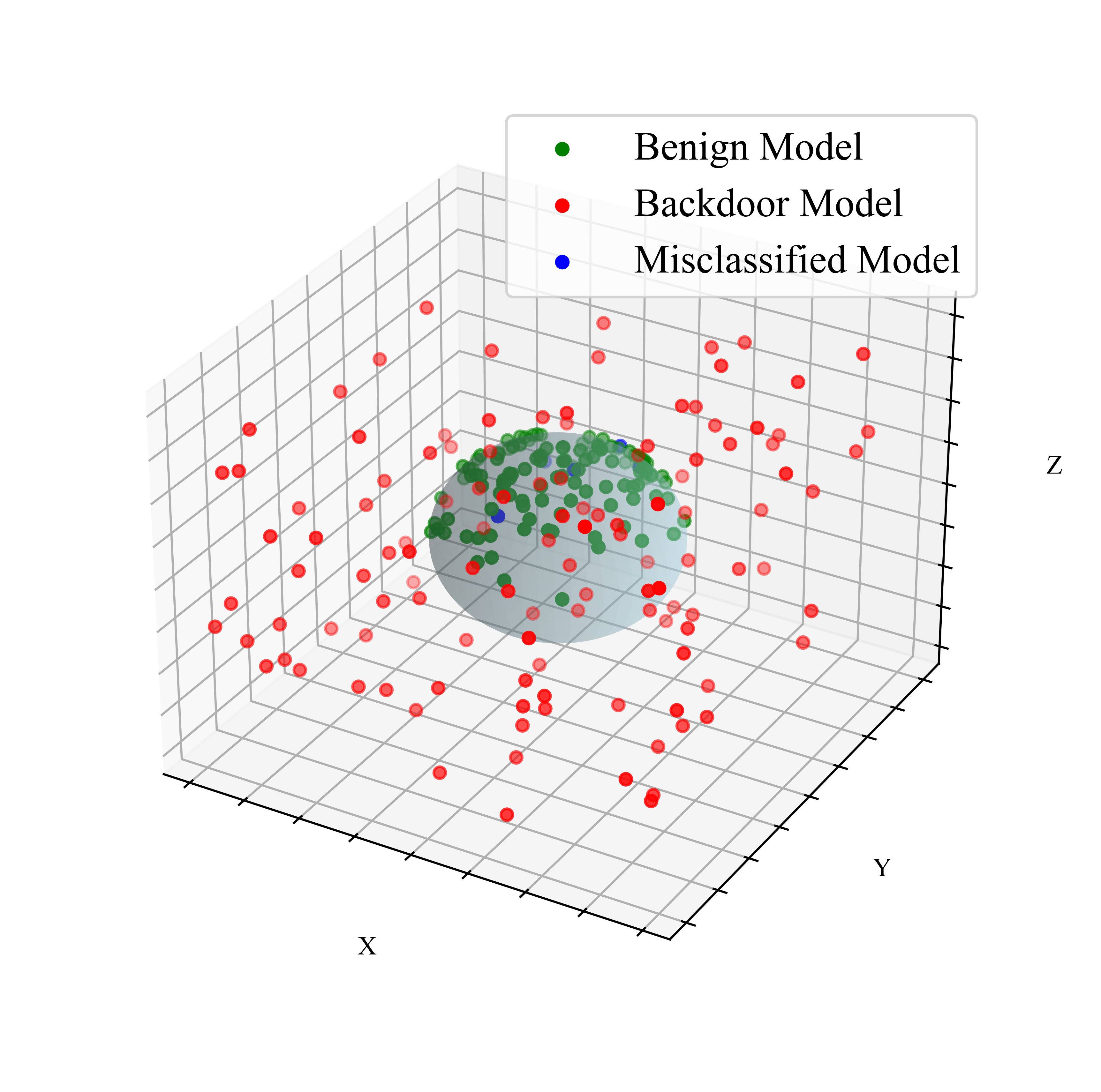}
    \caption{CIFAR10}
    \label{occ_cifar}
  \end{subfigure}
  \begin{subfigure}[b]{0.22\textwidth}
    \includegraphics[width=\textwidth]{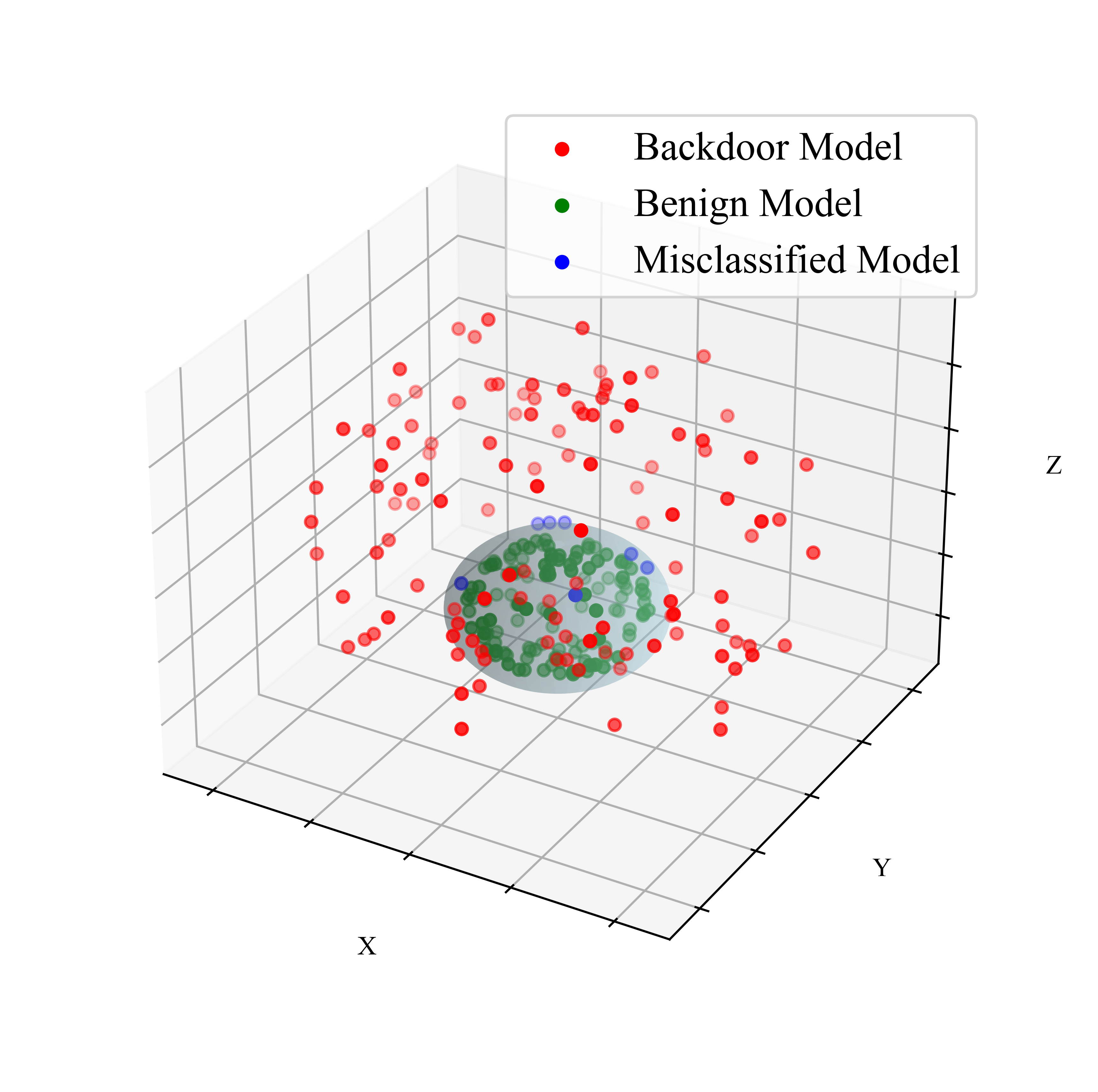}
    \caption{SC}
    \label{occ_SC}
  \end{subfigure}

\caption{Green dots denote benign models, red for backdoor, blue for misclassified, encircled by the one-class classifier's hypersphere.}
  \label{occ_view}
\end{figure}

Our OCGEC method maintains an average detection AUC of more than 90\% for all types of backdoor attacks on a variety of tasks when trained with only benign samples and has outstanding detection AUCs over 99\% for some attacks. OCGEC's AUC performance on the GTSRB dataset was approximately 3\% to 30\% higher than the other baselines (ours: 88.19\% lower than NC-B: 89.84\%), 5\% to 40\% higher on CIFAR10 (ours: 85.94\% lower than ABD-B: 88.28\%), 10\% to 30\% higher on SC (ours: 91.41\%  lower than NC: 96.48\% ), 5\% to 10\% higher on MR (ours: 93.75\% lower than Spectral-M: 95.31\%).
Further, we set the target label of the backdoor as  All-to-All attack based on the Modification Attack, and evaluate the detection performance on GTSRB and CIFAR10. The results are shown in Table  \ref{AlltoAll}. Our OCGEC model also outperforms the other six methods. As illustrated in Figure \ref{occ_view}, it is easier to detect backdoors at the model level due to the well-trained one-class classifier creating a hypersphere that encompasses all benign samples while excluding backdoor samples from the subinterface.

\subsubsection{Robustness} 
To validate the impact of model architecture on detection efficacy and evaluate the generalizability of our OCGEC, we implemented it across six distinct architectures on the CIFAR10 dataset: ResNet-18, ResNet-50, DenseNet-121, DenseNet-169, MobileNet v2, and WRN-22-6. For these intricate networks, we incrementally enlarged the training dataset. We use 10\% of the dataset to train benign tiny models and 50\% for generating backdoor tiny models. We generate 200 benign tiny models converted into graphs for training OCGEC, and the test set consists of 25 benign and 25 backdoor tiny models. The attack is kept as Modification attack. The experimental results are shown in the table\ref{structure}, and the AUC of the tests is close to 90\% under each network structure, indicating that our method can be generalized to most network structures and has universal application.
\begin{table}[t]
  \centering
  \caption{The detection AUC of CIFAR10 on six different network architectures.}
    \begin{tabular}{ccc}
    \toprule
    \multicolumn{1}{c}{ResNet-18} & \multicolumn{1}{c}{ResNet-50} & \multicolumn{1}{c}{DenseNet-121} \\
    92.96 & 87.52 & 90.24 \\
    \hline\hline\noalign{\smallskip}	
    \multicolumn{1}{c}{DenseNet-169} & \multicolumn{1}{c}{MobileNet v2} & \multicolumn{1}{c}{WRN-22-6} \\
    86.56 & 90.72 & 94.88 \\
    \bottomrule
    \end{tabular}%
  \label{structure}%
\end{table}%
\subsubsection{Effectiveness On Adaptive Backdoor Attacks} 
This experiment investigates the OCGEC's capability to identify adaptive attacks. Here, the adversary has full access to OCGEC's parameters and its graph-based model representation. For example, the training loss of the original OCGEC is set to $L_{train}$, and $L_{trojaned}$ is the output when the model is detected as a backdoor model by OCGEC. An attacker can evade detection by minimizing $L_{trojaned}$ during training as follows:
\begin{equation}
    \begin{gathered}
\min _{\substack{\theta,  \boldsymbol{c} \\
X=\left\{\mathbf{X}_1, \ldots, 
\mathbf{X_{\mathbf{k}}}\right\}}} \lambda\cdot L_{train} + (1-\lambda)\cdot L_{trojaned}
\end{gathered}
\end{equation}
We explore OCGEC's robustness against two adaptive attacks, guided by $\lambda$ to balance detection precision and false positives. The first attack aligns neural activations of benign and triggered inputs, while the second caps the activation disparity. By fine-tuning, we craft subtle backdoor models, assessing OCGEC against these strategies as shown in Table \ref{adaptive}. Despite these attacks mimicking innocuous behavior, OCGEC's hyperspherical boundary effectively distinguishes them, maintaining an AUC mean above 80\%, albeit slightly lower than against standard attacks.

\begin{figure}[hbp]   
	
\centering
	
\begin{subfigure}[b]{0.24\textwidth}
    \includegraphics[width=\textwidth]{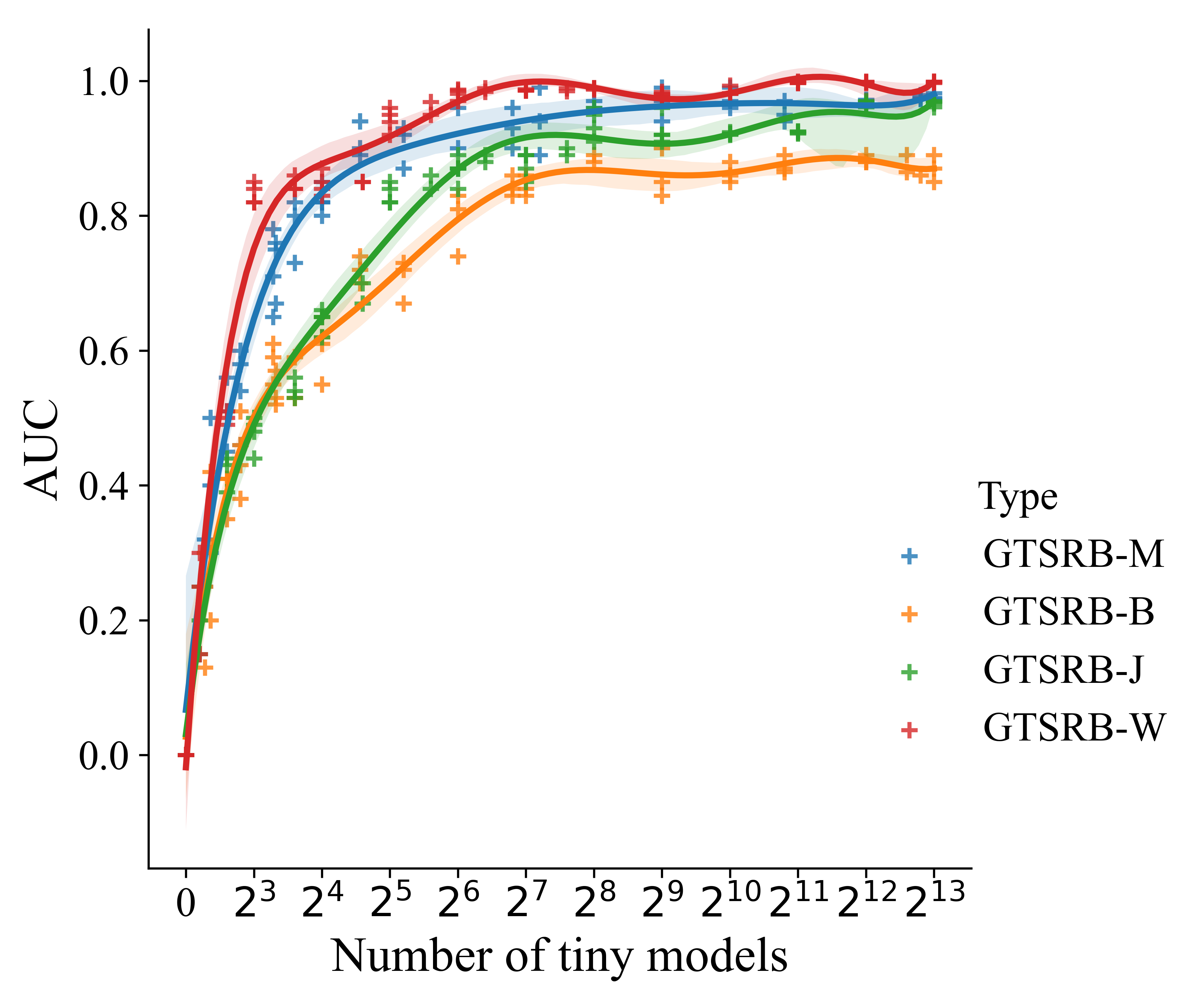}
    \caption{GTSRB}
    \label{number_tiny_GTSRB}
  \end{subfigure}
  \begin{subfigure}[b]{0.24\textwidth}
    \includegraphics[width=\textwidth]{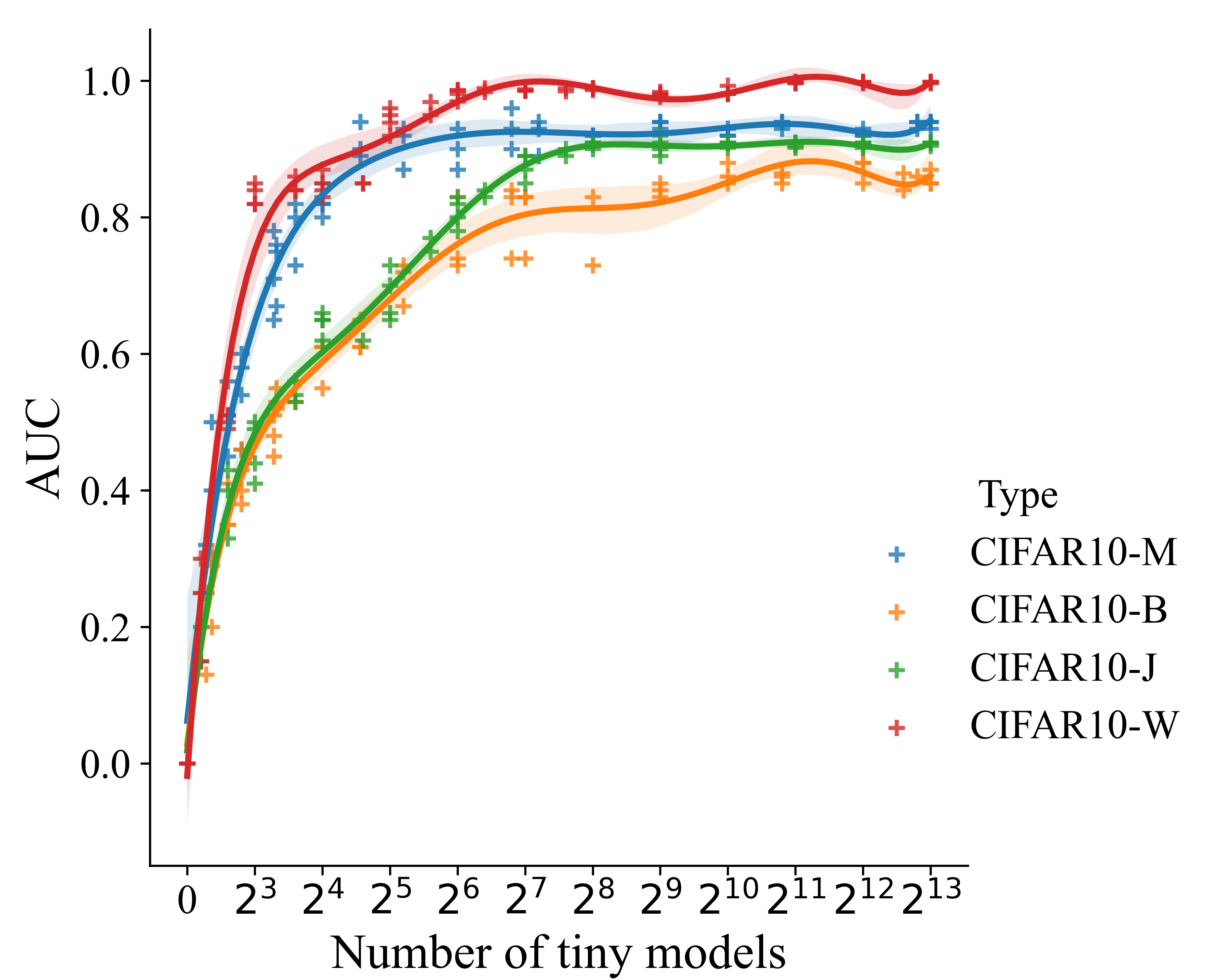}
    \caption{CIFAR10}
    \label{number_tiny_cifar}
  \end{subfigure}

\caption{Results of detection AUC with respect to the number of tiny models on GTSRB (a) and CIFAR10 (b) datasets.}
  \label{number}
\end{figure}
\subsection{Ablation Studies}
\noindent \textbf{Ablation Study On the Number of Tiny Models.} \
The sensitivity of OCGEC's detection performance on GTSRB and CIFAR10 with respect to the number of tiny models is illustrated in Figure \ref{number}. Our OCGEC achieves good detection results even with a small number of tiny models converted into graphs as the training set, and the detection performance continues to improve as the number of tiny models increases. There is a trade-off between the number of tiny models and the running time. To ensure a realistic usage scenario, the detector can take this situation into consideration.

\begin{figure}[htbp]   
	
\centering
	
\begin{subfigure}[l]{0.47\textwidth}
    \includegraphics[width=\textwidth,height=0.35\textwidth]{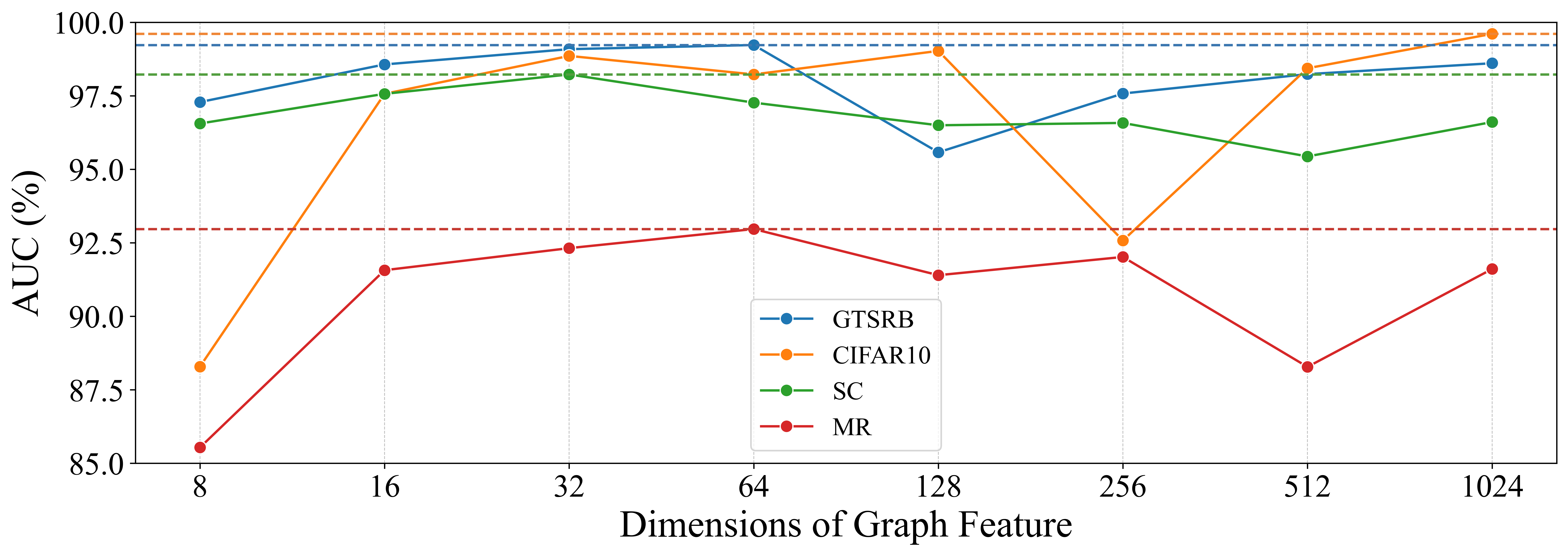}
    \caption{}
    \label{Ablation_dim}

  \end{subfigure}
  \begin{subfigure}[l]{0.47\textwidth}
    \includegraphics[width=\textwidth,height=0.35\textwidth]{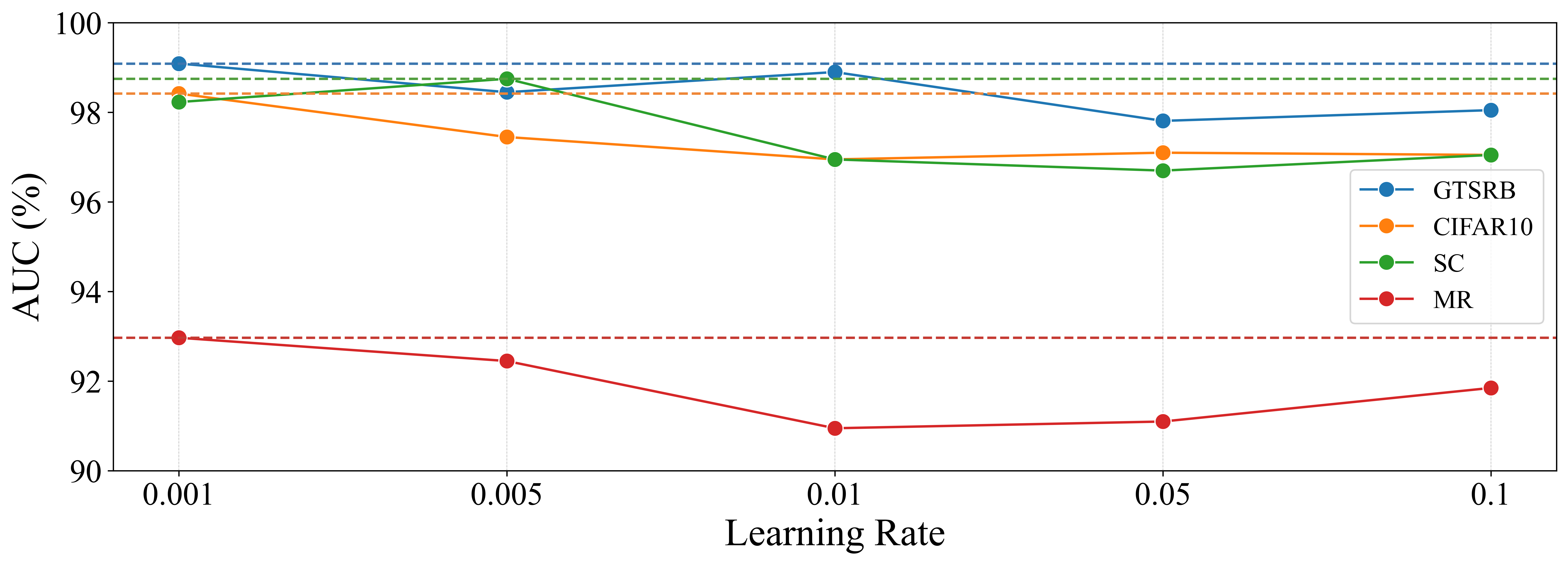}
    \caption{}
    \label{Ablation_LR}
  \end{subfigure}
\caption{Ablation studies with Jumbo Attack. We change the graph feature dimension (a) and learning rate (b) to explore the AUC performance of OCGEC.}
  \label{Ablation}
\end{figure}

\noindent \textbf{Ablation Study On Different  Graph Feature Dimensions.} \
The success of our backdoor detection is inextricably linked to the powerful feature extraction capabilities of GAE. Our one-class classification algorithm still has very good performance on high-dimensional representations. The impact on the performance of OCGEC is explored by adjusting the feature dimension of GAE encoding from a great span of 8 to 1024. The results are shown in Figure \ref{Ablation}(b). Our approach achieves almost the same performance. It shows that the single classification algorithm of OCGEC can also handle high-dimensional sparse features.

\noindent \textbf{Ablation Study On Different Learning Rates.} \ 
Figure \ref{Ablation}(a) shows that OCGCE is insensitive to the learning rate setting, suggesting that specialized settings are unnecessary for various backdoor attacks or task-specific datasets. The learning rate range was set from 0.001 to 0.1. Both the learning rate of GAE and the learning rate of the one-class classifier were adjusted simultaneously, as depicted in Figure \ref{Ablation}(b). The detection performance of OCGEC is consistent throughout a broad range of learning rates due to our robust one-class classification algorithm and joint optimization of GAE.

\noindent \textbf{Ablation Study On Different Feature Extraction Structure.} \ 
In this part, we explore the effectiveness of different graph feature extraction structures on the performance of the detector. To demonstrate the feature extraction effect of our masked GAE, we successively use PCA\cite{roweis1997algorithms}, VGAE\cite{kipf2016variational}, DeepWalk\cite{perozzi2014deepwalk}, the original GraphMAE\cite{houGraphMAESelfSupervisedMasked2022}, and our improved GAE for feature extraction. The results in Table \ref{FES} indicate that in large and sparse graphs, PCA no longer learns the principal component analysis in features, and VGAE and DeepWalk perform poorly. Our masked GAE is best suited for the downstream task of backdoor detection.
\begin{table}[htbp]
  \centering
  \caption{The detection AUC on two self-designed adaptive attacks.}
    \begin{tabular}{cccc}
    \toprule
    \multicolumn{1}{c}{GTSRB-No.1} & \multicolumn{1}{c}{GTSRB-No.2} & \multicolumn{1}{c}{CIFAR10-No.1}&\multicolumn{1}{c}{CIFAR10-No.2} \\
    87.54 & 91.88 & 83.39 & 85.81 \\
    \hline\hline\noalign{\smallskip}	
    \multicolumn{1}{c}{SC-No.1} & \multicolumn{1}{c}{SC-No.2} & \multicolumn{1}{c}{MR-No.1}&\multicolumn{1}{c}{MR-No.2} \\
    75.43 & 88.86 & 93.78 & 96.54  \\
    \bottomrule
    \end{tabular}%
  \label{adaptive}%
\end{table}%

\begin{table}[htbp]
  \centering
  \caption{Effectiveness of the Feature Extraction Structure on CIFAR10, SC, and MR datasets under Modification Attack. Evaluation metric: AUC in \%.}
    \begin{tabular}{cccc}
    \toprule
    Methods & \multicolumn{1}{c}{CIFAR10} & \multicolumn{1}{c}{SC} & \multicolumn{1}{c}{MR}\\
    \midrule
    PCA\cite{roweis1997algorithms}+OCC    & $\leq50$ & $\leq50$ & $\leq50$\\
    VGAE\cite{kipf2016variational}+OCC    & 60.67 & 57.44& 72.21\\
    DeepWalk \cite{perozzi2014deepwalk}+OCC   & 73.82 & 69.65 & 75.37\\
    GraphMAE\cite{houGraphMAESelfSupervisedMasked2022}+OCC & 84.36 & 90.16& 87.92 \\
\textbf{OCGEC} & \textbf{93.36} & \textbf{94.53} &\textbf{93.75}\\
    \bottomrule
    \end{tabular}%
  \label{FES}%
\end{table}%
\section{Conclusions}
In this paper, we propose One-class Graph Embedding Classification (OCGEC), a one-class classification framework that utilizes GNNs for model backdoor detection. OCGEC aims to convert model architecture and
weights features into graph data and then exploit the powerful representational capabilities of GNNs to map feature nodes to the hyperspheres in the embedding space. Our extensive experimental results show that the proposed OCGEC achieves superior performance compared to other backdoor detection methods with only small, clean data. Moreover, we have pioneered the idea of converting model-level backdoor features into graphs, thus achieving unprecedented generality in the field of model-level detection. We hope that our work will spur more comprehensive backdoor detection efforts, which is a prerequisite for backdoor defense.

\section*{Acknowledgment}
This work was supported in part by the National Key R\&D Program
of China under Grant 2023YFB3106502.


\bibliographystyle{IEEEtran}
\bibliography{IEEEabrv,GNN}

\end{document}